\documentclass[11pt]{article}

\usepackage[preprint]{acl}

\usepackage{times}
\usepackage{latexsym}

\usepackage[T1]{fontenc}

\usepackage[utf8]{inputenc}

\usepackage{microtype}

\usepackage{inconsolata}

\usepackage{graphicx}
\usepackage{hyperref}
\usepackage{url}
\usepackage{algorithm}
\usepackage{algorithmic}
\usepackage{booktabs} 
\usepackage{xcolor}
\usepackage{bbding}
\usepackage{longtable}
\usepackage{array}
\usepackage[many]{tcolorbox}
\usepackage{enumitem}
\usepackage{caption}
\usepackage{subcaption}
\usepackage{wrapfig}
\usepackage{amsmath}
\usepackage{amssymb}   
\usepackage{amsfonts}  

\title{OxyGent: Making Multi-Agent Systems Modular, Observable, and Evolvable via Oxy Abstraction}

\author{
  Junxing Hu\thanks{Equal contribution.} \quad Tianlong Li\footnotemark[1] \quad Lei Yu \quad Ai Han\thanks{Corresponding author.} \\
  JD.com \\
  Beijing, China \\
  \texttt{junxing.hu@cripac.ia.ac.cn, \{litianlong26, yulei116, hanai5\}@jd.com}
}

\begin{document}
\maketitle


\begin{abstract}

Deploying production-ready multi-agent systems (MAS) in complex industrial environments remains challenging due to limitations in scalability, observability, and autonomous evolution. We present OxyGent, an open-source framework driven by two core novelties: a unified Oxy abstraction and the OxyBank evolution engine. The unified abstraction encapsulates agents, tools, LLMs, and reasoning flows as pluggable atomic components, enabling Lego-like scalable system composition and non-intrusive monitoring. To enhance observability, OxyGent introduces permission-driven dynamic planning that replaces rigid workflows with execution graphs generated at runtime, providing adaptive visualizations. Furthermore, to support continuous evolution, OxyBank serves as an AI asset management platform that drives automated data backflow, annotation, and joint evolution. Empirical evaluations and real-world case studies show that OxyGent provides a robust and scalable foundation for MAS. OxyGent is fully open-sourced under the Apache License 2.0 at \url{https://github.com/jd-opensource/OxyGent}.

\end{abstract}

\section{Introduction}

Large Language Model (LLM) agents are transitioning from experimental prototypes to collaborative Multi-Agent Systems (MAS) capable of solving sophisticated real-world tasks~\cite{wang2024survey}. However, in large-scale distributed environments such as intelligent assistants~\cite{fu2024camphor}, existing MAS frameworks encounter critical bottlenecks. Traditional development paradigms may lack standardized abstractions, leading to low efficiency and repetitive implementation of core functions. 
Moreover, rigid plan-and-execute workflows are too brittle to handle the uncertainties of dynamic environments. 
In addition, without a closed-loop mechanism to evaluate and refine agents using online feedback, performance tends to stagnate or even deteriorate~\cite{gao2025survey}.

\begin{figure}[t]
\centering
\includegraphics[width=0.48\textwidth]{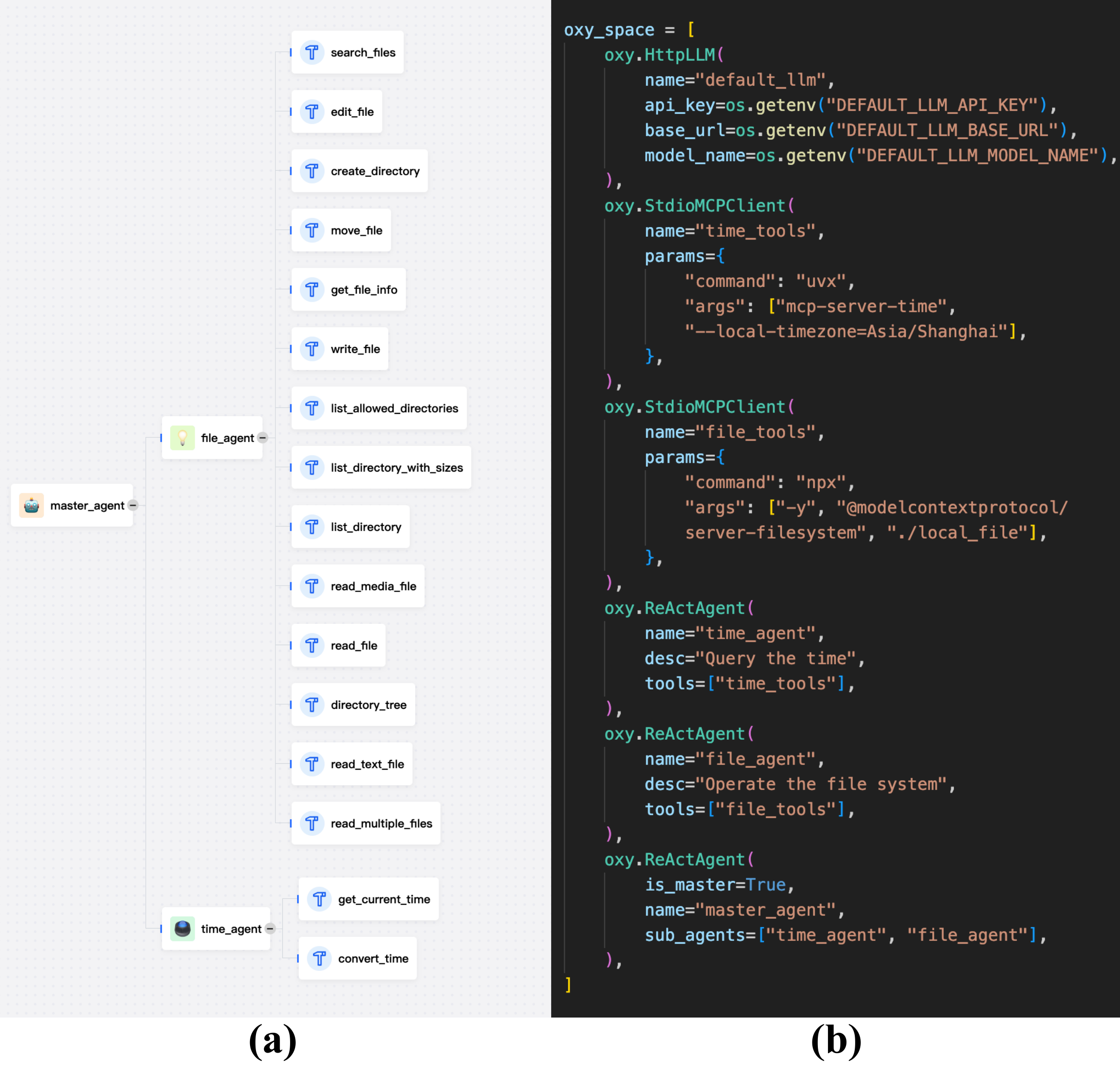}
\caption{
A multi-agent file management assistant built with OxyGent. (a) MAS visualization. (b) Permission relationships are defined in the implementation code.
}
\vspace{-3mm}
\label{oxy_sample}
\end{figure}



To enhance scalability and development efficiency, OxyGent introduces a unified abstraction that encapsulates agents, tools, and reasoning flows as pluggable Oxy nodes. This Lego-like approach empowers researchers to assemble and hot-swap specialized components without complex manual reconfiguration rapidly. Furthermore, by integrating Aspect-Oriented Programming (AOP)~\cite{kiczales1997aspect}, the framework separates core business logic from cross-cutting concerns, maintaining a clean, modular architecture for seamless expansion in production settings. Building on this abstraction to improve observability, OxyGent replaces static DAGs with permission-driven, dynamic planning. By defining potential collaboration spaces through node permissions (Figure~\ref{oxy_sample}(b)), the system automatically synthesizes execution paths at runtime and generates real-time call graph visualizations (Figure~\ref{oxy_sample}(a)). This transparency allows users to thoroughly inspect decision-making trajectories, ensuring that emergent MAS behaviors remain fully monitorable and manageable even in volatile task environments.

To facilitate the agent evolution, OxyGent introduces OxyBank, a specialized platform for AI asset management. OxyBank serves as the evolutionary engine of OxyGent by providing a systematic pipeline that captures online execution traces and converts them into high-quality training samples. 
By integrating automated rewarding and human-in-the-loop annotation, OxyBank enables the backflow of domain-specific knowledge into the multi-agent system. This closed-loop process supports the joint evolution of agent strategies, ensuring that the collective intelligence of the ecosystem continuously improves through data-centric learning rather than remaining static after deployment.

While many recent frameworks emphasize engineering integration, OxyGent's central research novelties lie in two foundational pillars: the unified Oxy abstraction and the OxyBank evolution engine. The main contributions are summarized as follows:

\begin{itemize}[leftmargin=*]

\item \textbf{Unified Oxy Abstraction.} 
We propose the unified Oxy abstraction to resolve structural heterogeneity in MAS, enabling Lego-like construction and high scalability.

\item \textbf{OxyBank Evolution Engine.} We introduce an AI asset platform that drives continuous MAS self-improvement via automated data backflow and verifiable evaluation.

\item \textbf{System Observability.} Built upon the Oxy abstraction, we implement permission-driven dynamic planning and an AOP-infused orchestration paradigm, ensuring deep observability.

\item \textbf{Empirical Validation \& Open Source.} Evaluations demonstrate OxyGent's robustness. We release the complete framework, built-in toolkits, a trace visualization Web UI, and comprehensive reference examples.

\end{itemize}

\section{Related Work}

\textbf{Orchestration Paradigms and Dynamic Planning.}
Current MAS frameworks focus on balancing agent autonomy with structural control through diverse paradigms. LangGraph~\cite{langchain2025langgraph} serves as a low-level orchestration framework and runtime for building long-running, stateful agents via graph structures. CrewAI~\cite{crewAIInc2025crewai} utilizes a role-driven approach, combining the collaborative intelligence of ``Crews'' with the precise control of ``Flows'' to manage complex processes. Beyond graph and role-based designs, Semantic Kernel~\cite{microsoft2024semantic} integrates AI capabilities with enterprise business logic, while MetaGPT~\cite{hong2024metagpt} encodes Standardized Operating Procedures (SOPs) into prompt sequences to mimic software company workflows. Recent research has also explored more fluid architectures. Aime~\cite{shi2025aime} replaces fixed planners with dynamic actor instantiation, and AFlow~\cite{zhang2025aflow} automates the generation of agentic workflows. OxyGent distinguishes itself by unifying agents, tools, and flows into interchangeable atomic nodes governed by permission-driven dynamic planning and real-time execution visualization.

\textbf{Execution Lifecycle and System Observability.}
Observability is a critical production-readiness requirement, yet many frameworks prioritize developer experience over runtime monitoring. 
OpenAI Agents SDK~\cite{openai2025sdk}, the successor to Swarm~\cite{openai2024swarm}, adopts a minimalist set of primitives, including Agents, Handoffs, and Guardrails, and provides built-in tracing for debugging agentic flows.
AutoGen~\cite{wu2024autogen} models interactions as asynchronous message passing between entities, while Pydantic AI~\cite{pydantic2024pydantic} introduces a type-safe Python framework that enforces strict contracts for tool calling and data exchange. For enterprise-grade monitoring, Strands Agents~\cite{aws2025strands} provides a model-agnostic toolkit optimized for OpenTelemetry integration, similar to the durable execution principles in Dapr Agents~\cite{dapr2025dapr}. OxyGent implements a comprehensive AOP-based lifecycle management system that modularly injects cross-cutting concerns such as security auditing and performance monitoring into distributed agent joinpoints.

\begin{figure*}[t]
\centering
\includegraphics[width=1.0\textwidth]{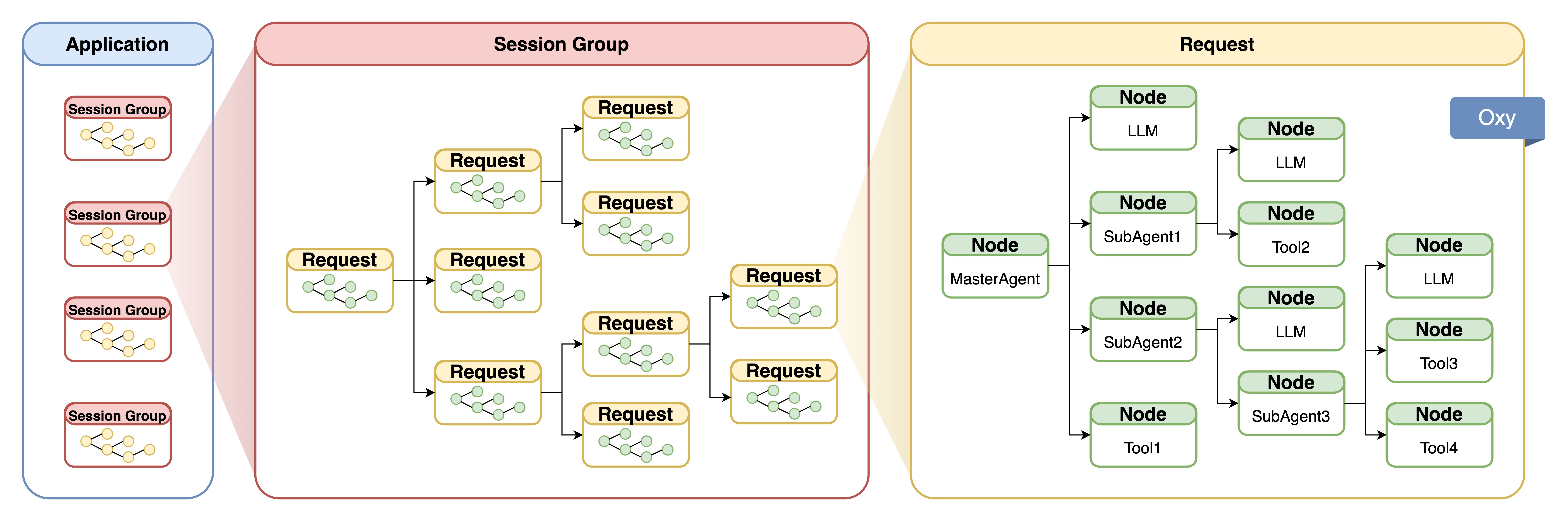}
\caption{
The OxyGent framework provides four data scopes: Application, Session Group, Request, and Node. 
This multi-level data isolation and sharing ensures data management efficiency and development convenience in MAS.
}
\vspace{-3mm}
\label{oxy_data}
\end{figure*}

\begin{figure*}[t]
\centering
\includegraphics[width=1.0\textwidth]{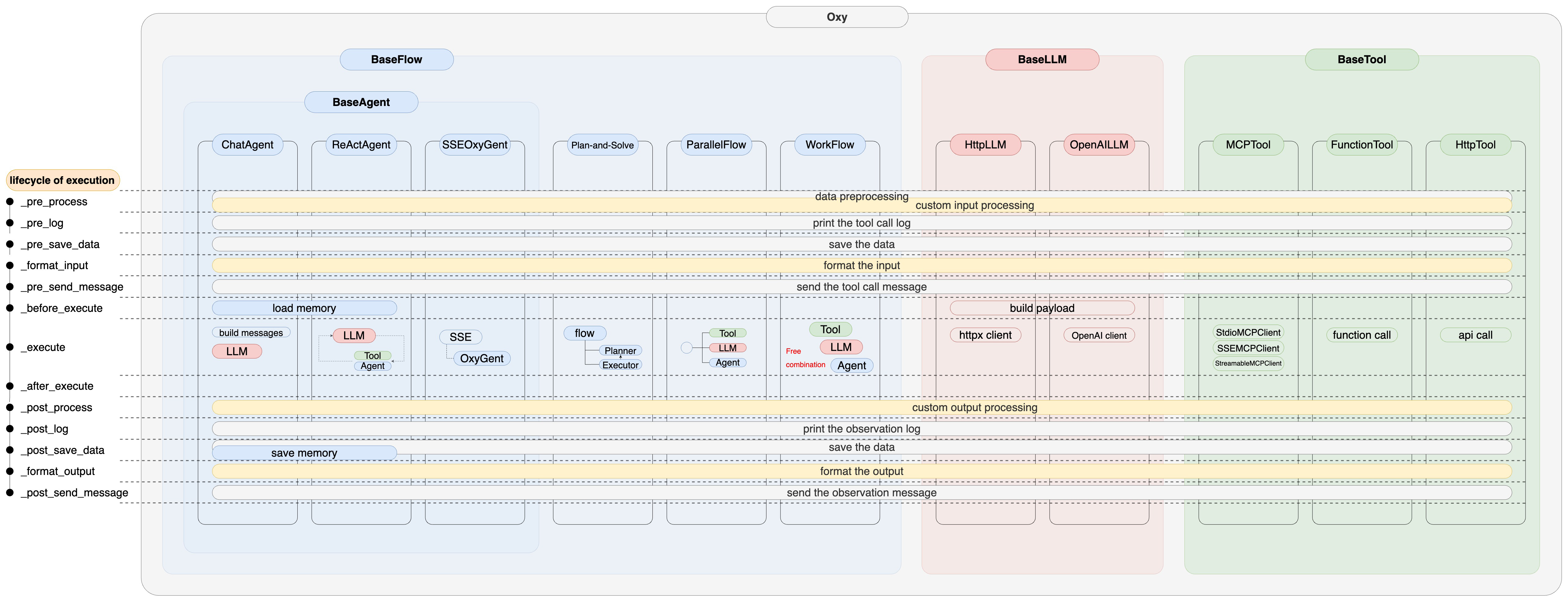}
\caption{
The execution lifecycle of Oxy. A series of management and coordination steps ensures data flow and processing, enabling MAS to dynamically plan and generate flowcharts in real time.
}
\vspace{-3mm}
\label{oxy_process}
\end{figure*}

\textbf{Evolutionary AI Asset Management.}
The transition toward agents that learn and improve has led to systems that prioritize memory and training loops~\cite{han2025joyagents}. Agno~\cite{agno2025agno} serves as specialized infrastructure for reasoning systems, utilizing its AgentOS layer to manage long-term memory and knowledge accumulation across sessions. AWorld~\cite{yu2025aworld} provides a high-performance interaction runtime designed for large-scale experience generation to enable practical reinforcement learning for agents. EvoAgent~\cite{yuan2025evoagent} and SE-Agent~\cite{lin2025se} have proposed automatic multi-agent extension via evolutionary operators and cross-trajectory refinement, respectively. 
Moreover, OWL~\cite{hu2025owl} improves cross-domain generalization by optimizing a domain-agnostic planner with reinforcement learning from real-world feedback, whereas Chain-of-Agents~\cite{zhang2024chain} focuses on distilling multi-agent reasoning patterns into unified foundation models.
OxyGent facilitates a continuous build-inference-evolution cycle through OxyBank, an industrial-scale AI asset platform that unifies data annotation and knowledge backflow to improve collective agents.

\begin{figure*}[t]
\centering
\includegraphics[width=1.0\textwidth]{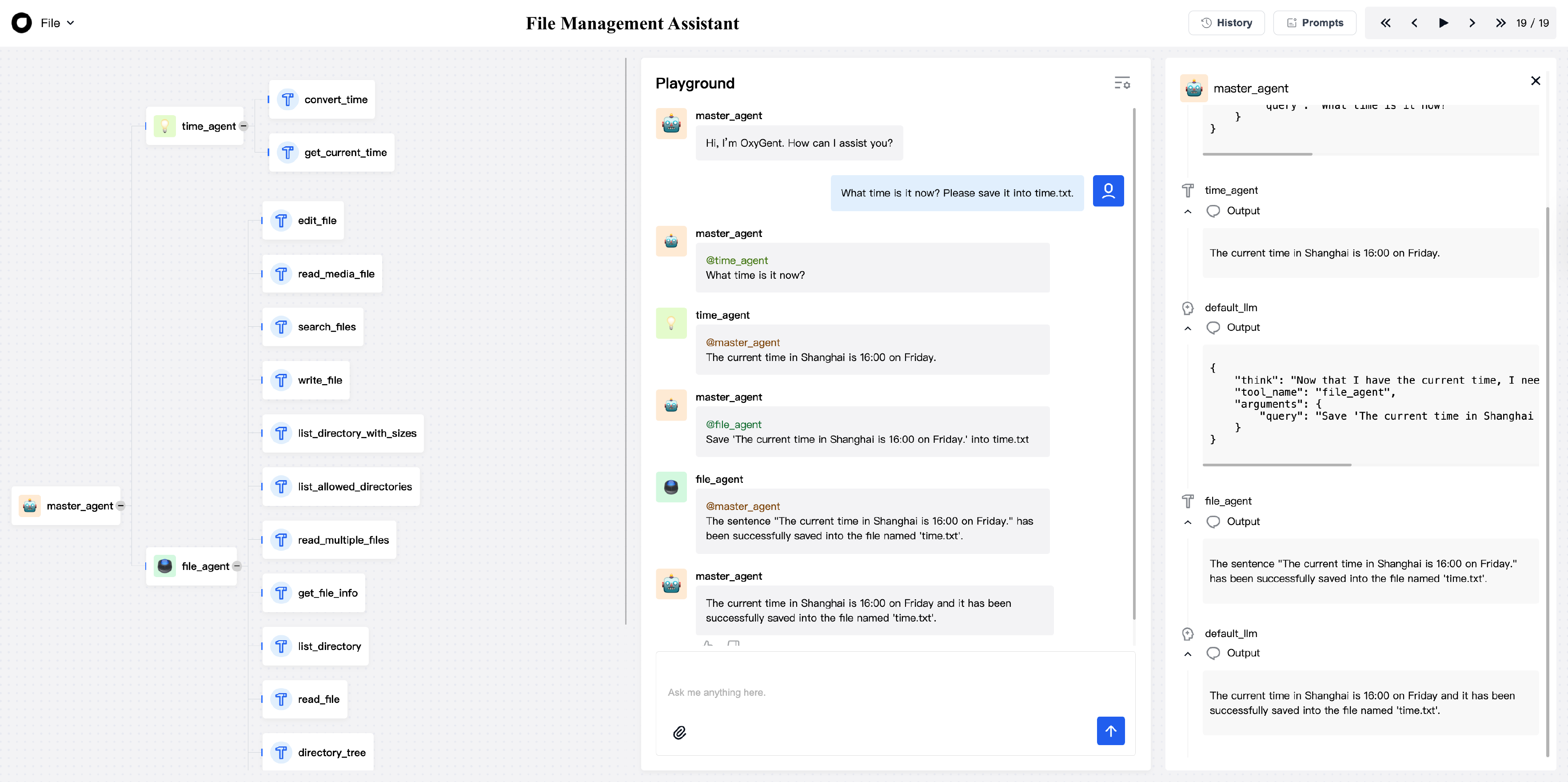}
\caption{
The file management assistant is built on OxyGent. From left to right: MAS visualization, question-answering box, and the output of each node. 
As shown in Figure~\ref{oxy_monitor_appendix}, each intermediate step of the reasoning process can be rolled back and reviewed.
}
\vspace{-3mm}
\label{oxygent_full_sample}
\end{figure*}

\begin{figure*}[t]
\centering
\includegraphics[width=1.0\textwidth]{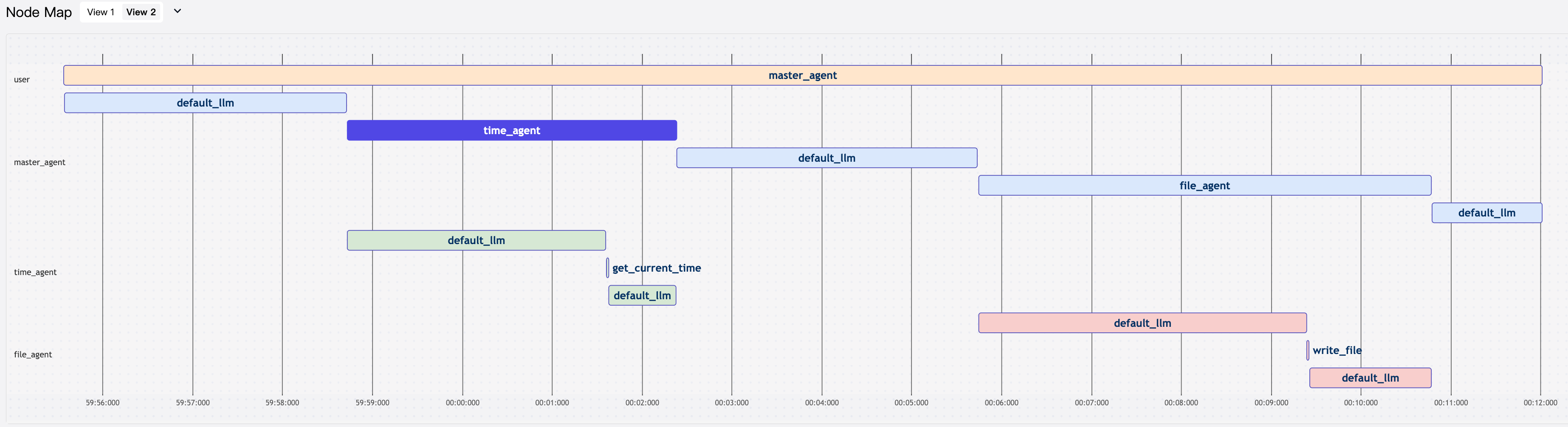}
\caption{
MAS inference monitoring. OxyGent has built-in production-grade time tracking, which displays task distribution and resource congestion in real time, facilitating MAS architecture optimization. 
More features are introduced in Appendix~\ref{appendix_features}.
}
\vspace{-3mm}
\label{oxy_monitor}
\end{figure*}

\begin{figure*}[t]
\centering
\includegraphics[width=1.0\textwidth]{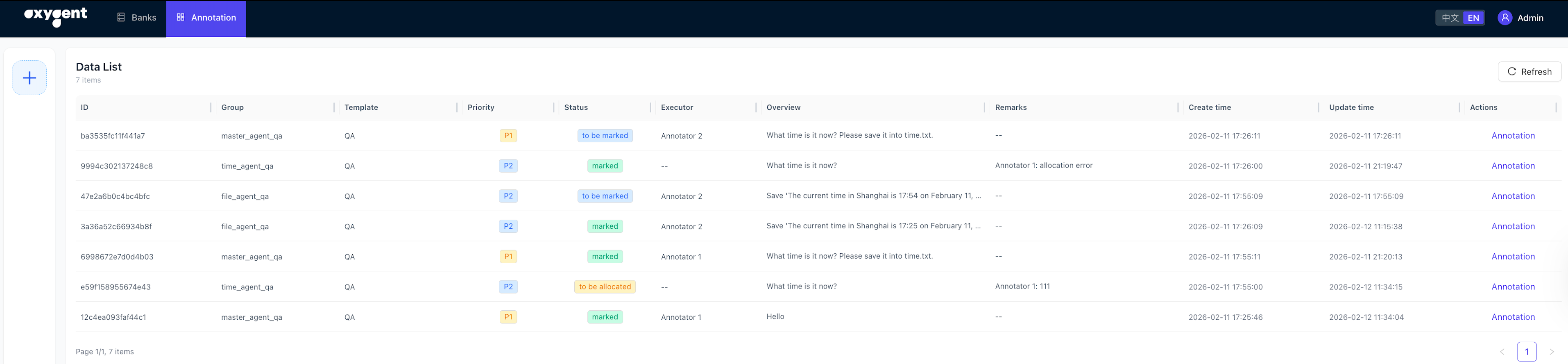}
\caption{
OxyBank, as OxyGent's one-stop AI asset management platform, supports knowledge base construction, data annotation, memory management, multi-agent system evaluation, and enables agent evolution.
}
\vspace{-3mm}
\label{oxy_bank}
\end{figure*}

\section{OxyGent}

OxyGent aims to transform multi-agent orchestration from static, hard-coded workflows into an elastic, Lego-like assembly of atomic components. This section details the framework's infrastructure, the dynamic planning mechanisms, and the evolutionary loop powered by OxyBank.

\subsection{Unified Oxy Abstraction and Hierarchical Data Scopes}
To facilitate the flexible construction of complex MAS, {OxyGent} introduces a {unified atomic abstraction} where agents, tools, LLMs, and flows are wrapped into pluggable \textbf{Oxy} components. Unlike traditional frameworks that treat these entities as distinct software layers, {OxyGent} provides a consistent interface for all AI atoms, enabling hot-swapping of reasoning logic and capabilities at runtime without refactoring.

A cornerstone of unified abstraction is the four-tier data scoping mechanism, which provides a structured approach to state management in distributed environments (Figure~\ref{oxy_data}). Rather than relying on a single global state, {OxyGent} isolates data into specific domains to prevent data-privilege leaks and ensure efficiency. The domain description and data processing tools are as follows:

\begin{itemize}
    \item \textbf{Application}: Global context accessible across all MAS instances. 
    Data tools are: \texttt{oxy\_request.get/set\_global\_data()}  
    \item \textbf{Session Group}: Shared memory among a cluster of related conversations. 
    Data tools are: \texttt{oxy\_request.get/set\_group\_data()}  
    \item \textbf{Request}: Transient data restricted to a single inference trajectory. 
    Data tools are: \texttt{oxy\_request.get/set\_shared\_data()}  
    \item \textbf{Node}: Local arguments specific to an individual Oxy component. 
    Data tools are: \texttt{oxy\_request.get/set\_arguments()}  
\end{itemize}

Through hierarchical data isolation and sharing, the efficiency of data management and the ease of development in multi-agent collaborative scenarios are significantly improved.

\subsection{Permission-Driven Planning and Standardized Execution Lifecycle}
In contrast to fixed workflows, {OxyGent} utilizes {permission-driven dynamic planning} to govern agent collaboration. The execution trajectory is not predefined but emerges from the authorization relationships between Oxy nodes. To support this dynamism while maintaining observability, we implement a comprehensive Oxy execution lifecycle as shown in Figure~\ref{oxy_process}. Each node traverses a standardized sequence of management steps, and the example operations for each stage are as follows:

\begin{itemize}
    \item \textbf{Pre-execution Phase}: \texttt{\_pre\_process} is used for data formatting and \texttt{\_pre\_save\_data} persists inputs before logic invocation.
    \item \textbf{Core Execution}: The \texttt{\_execute} hook triggers the actual LLM reasoning or tool usage.
    \item \textbf{Post-execution Phase}: \texttt{\_post\_process} for data post-processing and \texttt{\_format\_output} for downstream compatibility.
\end{itemize}

By utilizing Aspect-Oriented Programming (AOP), {OxyGent} injects observability and security operators into these lifecycle joinpoints. This enables the system to automatically synthesize real-time execution visualizations of the actual call graphs during runtime (Figure~\ref{oxygent_full_sample}). 
Furthermore, as shown in Figure~\ref{oxy_monitor}, OxyGent incorporates production-grade time tracking to expose real-time agent-level resource consumption. By decomposing execution into LLM inference, tool and API calls, and inter-agent interactions, it enables efficient bottleneck identification and fine-grained architectural optimization.

\subsection{Evolutionary AI Asset Management via OxyBank}
The long-term value of an MAS lies in its ability to self-improve. We introduce OxyBank, an AI asset management platform serving as the framework's memory bank (Figure~\ref{oxy_bank}). It bridges the gap between raw execution data and continuous self-improvement through the following mechanisms:

\textbf{Closed-Loop Asset Pipeline.} 
The framework supports autonomous evolution by establishing a closed-loop cycle for annotation, auditing, and data backflow. Online execution traces are automatically captured and precipitated as memory assets within OxyBank. These assets are processed by annotation agents or experts using customized templates (\emph{e.g.}, QA or business-domain tagging). Once audited, the high-quality samples are used to update knowledge bases or fine-tune models. This pipeline ensures that the MAS continuously refines its decision-making strategies based on validated feedback from production.

\textbf{Multi-Layered Quality Gating.} 
To address potential degradation from noisy traces, OxyBank implements a strict multi-layered quality-gating pipeline. Specifically, it employs MD5-based deduplication at deposit to prevent frequency bias, and infers trace priorities based on call chain metadata (\emph{e.g.}, end-to-end user interactions are prioritized as P0). Crucially, no raw trace can bypass the strict state machine (pending $\rightarrow$ annotated $\rightarrow$ approved). The pending or rejected data are strictly isolated from the knowledge base, ensuring that only high-fidelity signals drive the evolution process.

\begin{figure}[t]
\centering
\includegraphics[width=1.0\linewidth]{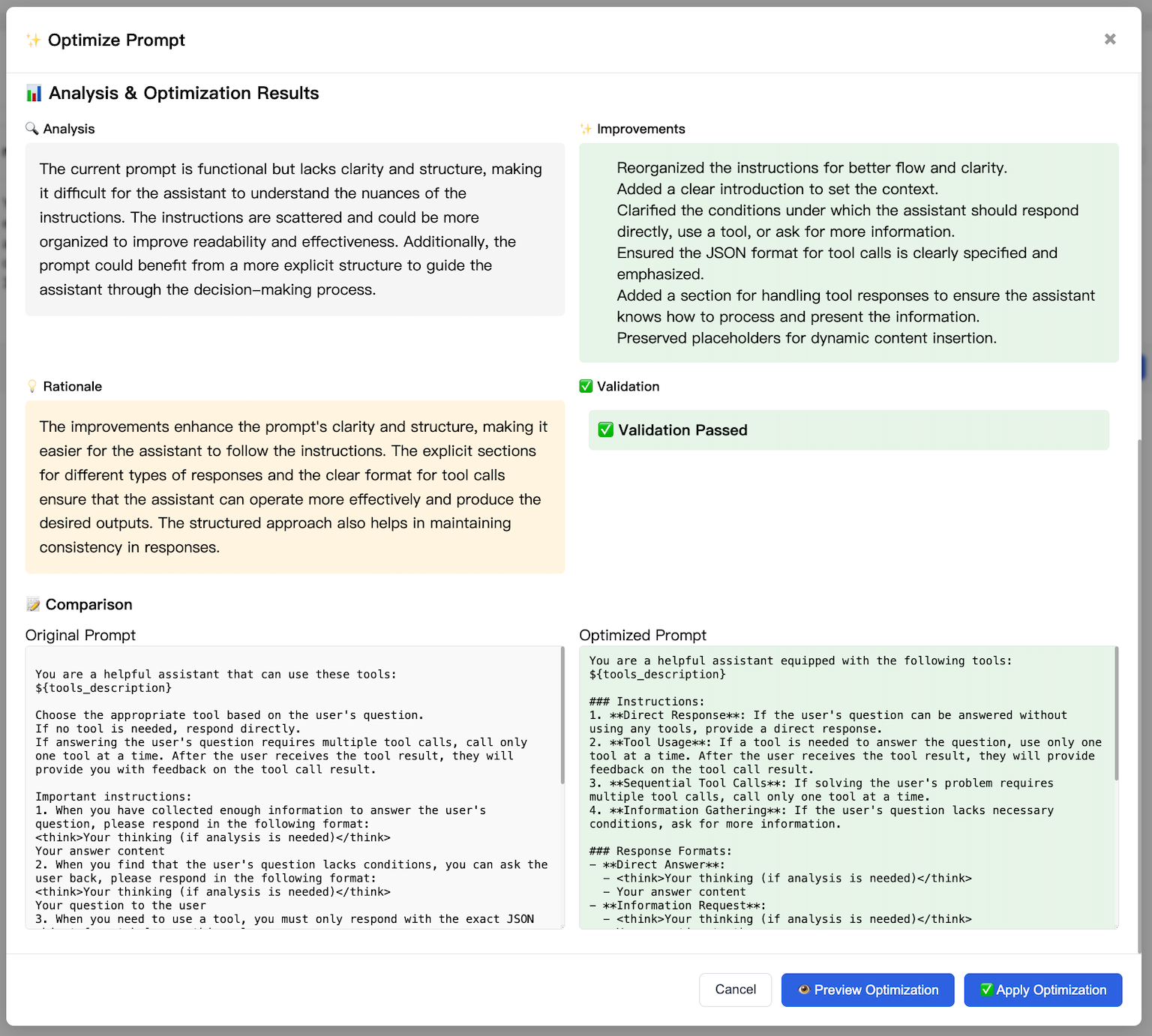}
\caption{
AI-driven Optimize Prompt module in OxyGent, which automatically analyzes and refines agent prompts based on historical execution traces, improving prompt quality without manual engineering.
}
\vspace{-3mm}
\label{prompt_optimize}
\end{figure}

\textbf{Autonomous Prompt Optimization.} 
Instead of relying solely on manual prompt engineering, OxyGent integrates an AI-driven Optimize Prompt module as shown in Figure~\ref{prompt_optimize}. It automates the extraction of best practices from validated traces, allowing agents to iteratively refine their system prompts based on historical execution contexts. This dynamically blends the reliability of human-in-the-loop auditing with the scalability of autonomous LLM optimization, significantly accelerating the agents' adaptation to novel tasks.

\begin{figure*}[t]
\centering
\includegraphics[width=1.0\linewidth]{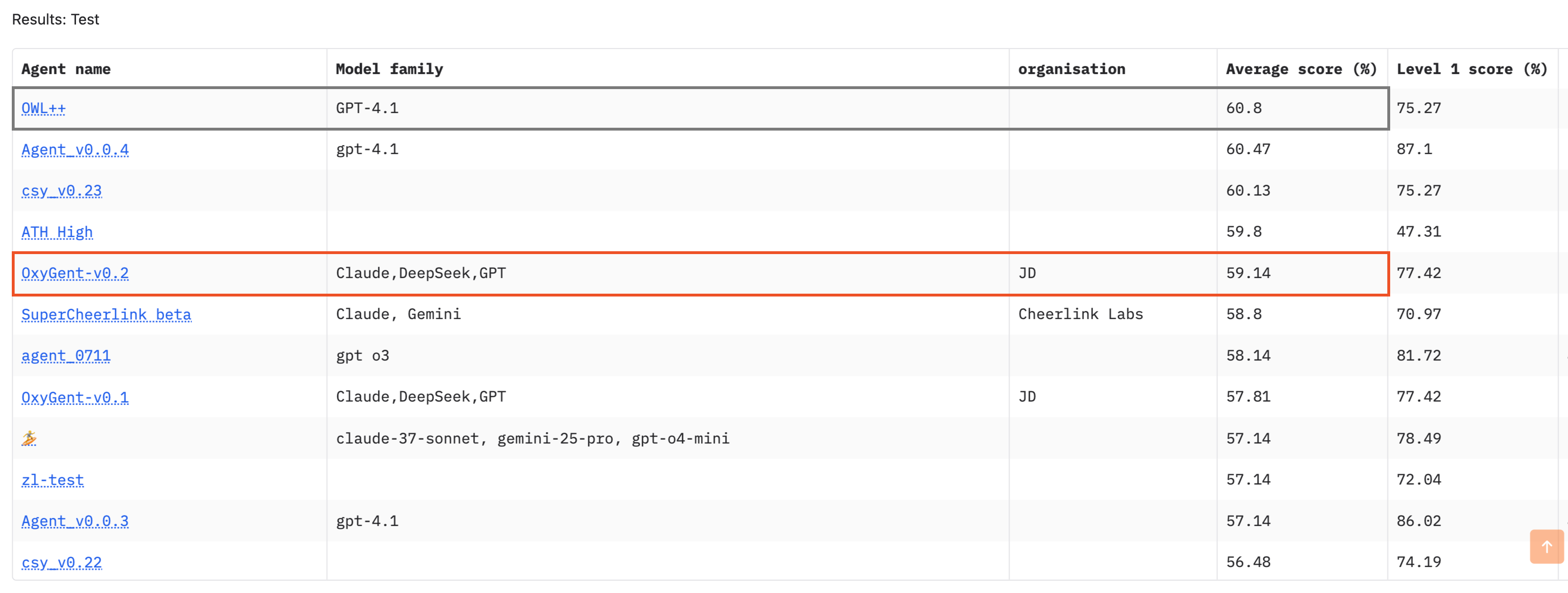}
\caption{
On July 22, 2025, OxyGent achieved the second-highest score (59.14\%) among open-source methods on the GAIA leaderboard, second only to OWL++, which was the strongest open-source method at the time.
}
\vspace{-3mm}
\label{case_gaia}
\end{figure*}

\section{Case Studies}

\subsection{Evaluations on the GAIA Benchmark }

\textbf{Benchmark.} GAIA~\cite{mialon2023gaia} focuses on real-world tasks that require fundamental human-level abilities. It is intuitive for humans, who achieve an average accuracy of 92\%, while frontier models like GPT-4, equipped with plugins, struggled at 15\%. The benchmark comprises 466 questions across 3 levels of increasing complexity.

\textbf{Experimental Results.} To evaluate the orchestration effectiveness of {OxyGent}, we tested the framework against the GAIA leaderboard. On July 22, 2025, {OxyGent} achieved a score of {59.14\%}, ranking as the \textbf{second-highest} open-source method globally at the time, closely trailing the open-source benchmark OWL++~\cite{hu2025owl} (60.8\%), as shown in Figure~\ref{case_gaia}. Although the emergence of newer state-of-the-art models and advanced reasoning toolsets has since moved the performance ceiling, these results highlight {OxyGent}'s inherent structural advantages in managing long-chain interactions and multi-modal dependencies efficiently. 
By leveraging its unified atomic abstraction, the framework enables high performance even with earlier-generation base models. 
The method is detailed in Appendix~\ref{appendix_gaia}.
The full implementation and specific configurations used for this evaluation are available for community review at: \url{https://github.com/jd-opensource/OxyGent/tree/gaia59}.

\begin{table}[t]
\centering
\small
\begin{tabular}{lcccc}
\toprule
Method & Avg & Level 1 & Level 2 & Level 3 \\
\midrule 
Single agent & 36.21  & 61.29  & 29.56  & 10.20  \\ 
+ Multi-agent & 42.19  & 62.37  & 35.85  & 24.49  \\ 
\quad + Planning & 52.16  & 62.37  & 54.09  & 26.53  \\ 
\quad \quad + Memory & \textbf{59.14}  & \textbf{77.42}  & \textbf{56.60}  & \textbf{32.65} \\
\bottomrule
\end{tabular}
\caption{
Ablation study on the GAIA benchmark (\%).
}
\label{gaia_ablation}
\end{table}

\textbf{Ablation Study.} To isolate the performance gains derived from our framework's specific designs, we conducted an ablation study on the GAIA benchmark (Table~\ref{gaia_ablation}). Since the unified Oxy abstraction serves as the non-removable foundation of the system, we evaluated the sequential addition of multi-agent orchestration, permission-driven dynamic planning, and the memory backflow (the predecessor to OxyBank). The results show that dynamic planning significantly boosts Level 2 reasoning (+18.24\%), while the memory mechanism predominantly enhances L1 task precision (+15.05\%) and overall average scores, validating that the gains stem from OxyGent's architectural designs rather than merely the underlying foundation models.

\begin{figure*}[t]
\centering
\includegraphics[width=1.0\linewidth]{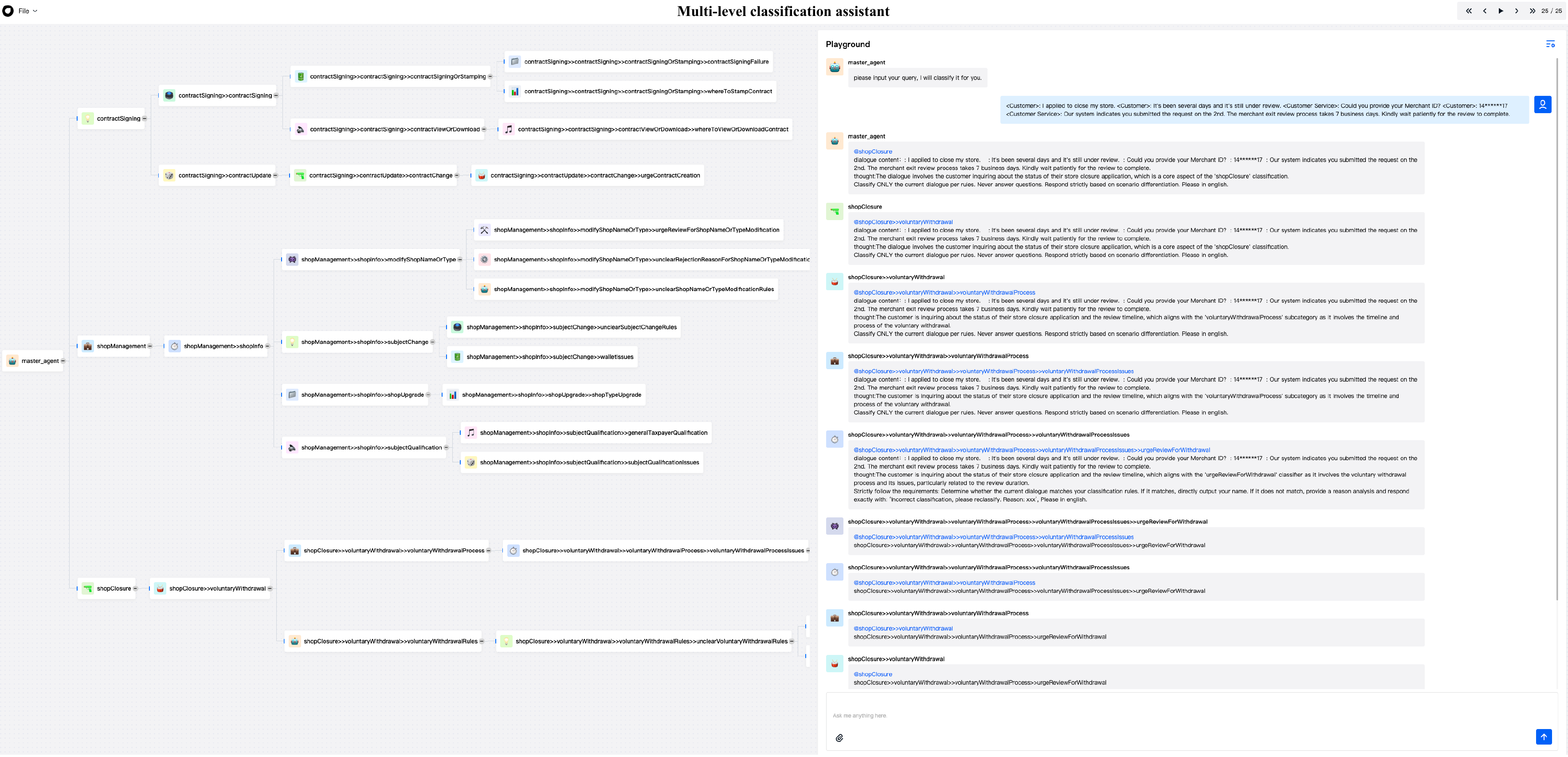}
\caption{
A hierarchical MAS of 2,000+ agents for e-commerce classification, employing a top-down decision chain with dynamic logic and full-chain tracing. ``>>'' denotes a call invocation between Oxy nodes.
}
\vspace{-3mm}
\label{case_2000agents}
\end{figure*}

\subsection{Case Analysis in Business Scenarios}

Beyond academic benchmarks, OxyGent has demonstrated significant industrial impact through wide-scale deployment in enterprise production and open-source communities. To illustrate the framework's capacity for extreme scale and complexity, we implemented a hierarchical multi-agent classification system for e-commerce customer service, as shown in Figure~\ref{case_2000agents}.

\textbf{Setup and Evaluation.} This system comprises over \textbf{2,000 agents} tasked with categorizing incoming requests into a taxonomy of more than 2,400 distinct labels. Each category contains as few as 10 samples, making the task a massive few-shot classification problem. We compared OxyGent against a strong single-agent baseline utilizing RAG retrieval coupled with 
DeepSeek-R1-Distill-Qwen-32B~\cite{guo2025deepseek}. The evaluation protocol continuously samples 2,000+ real-world interactions weekly for AI and human joint assessment.

\textbf{Results and Trade-offs.} The MAS-driven approach improved classification accuracy from 61.3\% (baseline) to 85.6\%. Furthermore, it demonstrated strong topological self-evolution, autonomously discovering and validating an average of 5.4 new categories per week. Regarding cost and latency tradeoffs, the dynamic generation of the 2,000+ agents keeps memory overhead manageable. While the multi-tiered decision architecture increases average inference latency by $2.3 \times$ compared to the single-agent baseline, this trade-off is strictly justified for offline business environments prioritizing high precision.

\textbf{Failure Cases.} We observed that highly ambiguous semantic queries occasionally induce repetitive ReAct loops or hallucinations, which are currently mitigated by our trace replay and manual auditing loops.
Due to space constraints, additional industrial scenarios including automated SOPs, RAG-based assistants, and community-contributed tools are detailed in Appendix~\ref{appendix_cases}.

\section{Discussion}

\textbf{Decision Routing and Structural Resilience.} To resolve execution conflicts when multiple nodes qualify, OxyGent utilizes a centralized orchestrator for dynamic, permission-constrained delegation. Under node failures, the framework prevents global trajectory collapse by encapsulating local errors into reasoning memory, enabling autonomous corrective replanning.

\textbf{Connections to Concurrent Works.} 
AlphaApollo~\cite{zhou2026alphaapollo} focuses on deep agentic reasoning and learning. In contrast, OxyGent provides a general-purpose, modular, and observable framework, designed for reliable industrial evolution via verifiable data backflow. For inspectability, Landscape of Thoughts~\cite{zhou2026landscape} provides post-hoc visual analysis of textual reasoning trajectories, whereas OxyGent visualizes multi-agent execution topologically at runtime for immediate architectural debugging. Finally, regarding AR-Bench~\cite{zhou2025passive}, while we evaluated on GAIA to test general multi-modal assistance, OxyGent's dynamic planning inherently supports the iterative information-seeking required by active reasoning tasks.

\section{Conclusion}

In this paper, we present OxyGent, an open-source framework for building scalable and self-improving multi-agent systems driven by two core novelties. First, its unified atomic abstraction enables a Lego-like construction process that replaces rigid workflows with permission-driven dynamic planning, offering high observability through real-time execution visualizations. Second, integrated with OxyBank for AI asset management, the framework continuously refines strategies via a validated feedback pipeline, as evidenced by evaluations on the GAIA benchmark and industrial applications.

\paragraph{Limitations.} 
Although OxyBank supports self-evolution, large-scale training currently depends on manual resource configuration. We are actively developing intelligent resource scheduling to achieve a fully automated lifecycle, including agent construction, deployment, and refinement. This will minimize human intervention and further accelerate the evolution of collective intelligence in complex environments.

\bibliography{custom}




\clearpage 
\onecolumn 

\newpage

\appendix

\section{Appendix}\label{appendix}

\subsection{More Features of OxyGent}\label{appendix_features}

Beyond the core orchestration and evolution pipelines, {OxyGent} provides a suite of tools designed for deep inspection and real-time intervention in AI decision-making.

\begin{figure*}[h]
\centering
\includegraphics[width=0.9\textwidth]{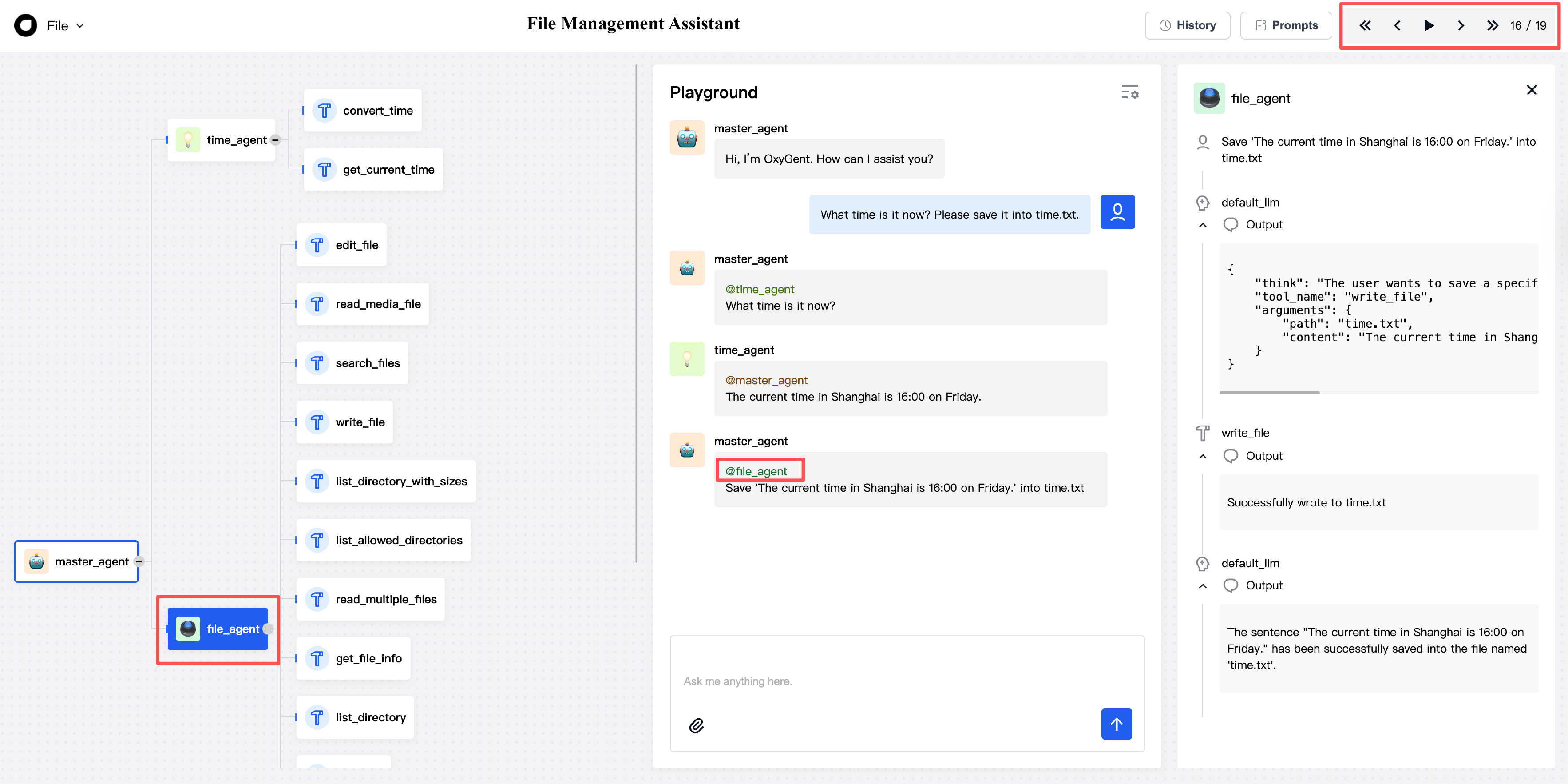}
\caption{
During runtime, the calling nodes on the left will be highlighted. When the inference is complete, any intermediate step can be retraced using the progress button in the upper right corner.
}
\vspace{-3mm}
\label{oxygent_full_sample_appendix}
\end{figure*}

\textbf{Interactive Decision Exploration.} OxyGent allows researchers to pause execution at any stage to analyze an agent's internal state, including LLM prompts, memory snapshots, and pending tool invocations. Developers can modify variables and re-sample inference paths on-the-fly without restarting the entire system. This enables parallel experimentation with different models or prompts, in which each variant is automatically recorded for seamless comparison and rollbacks, as shown in Figure~\ref{oxygent_full_sample_appendix}.

\begin{figure*}[h]
\centering
\includegraphics[width=0.9\textwidth]{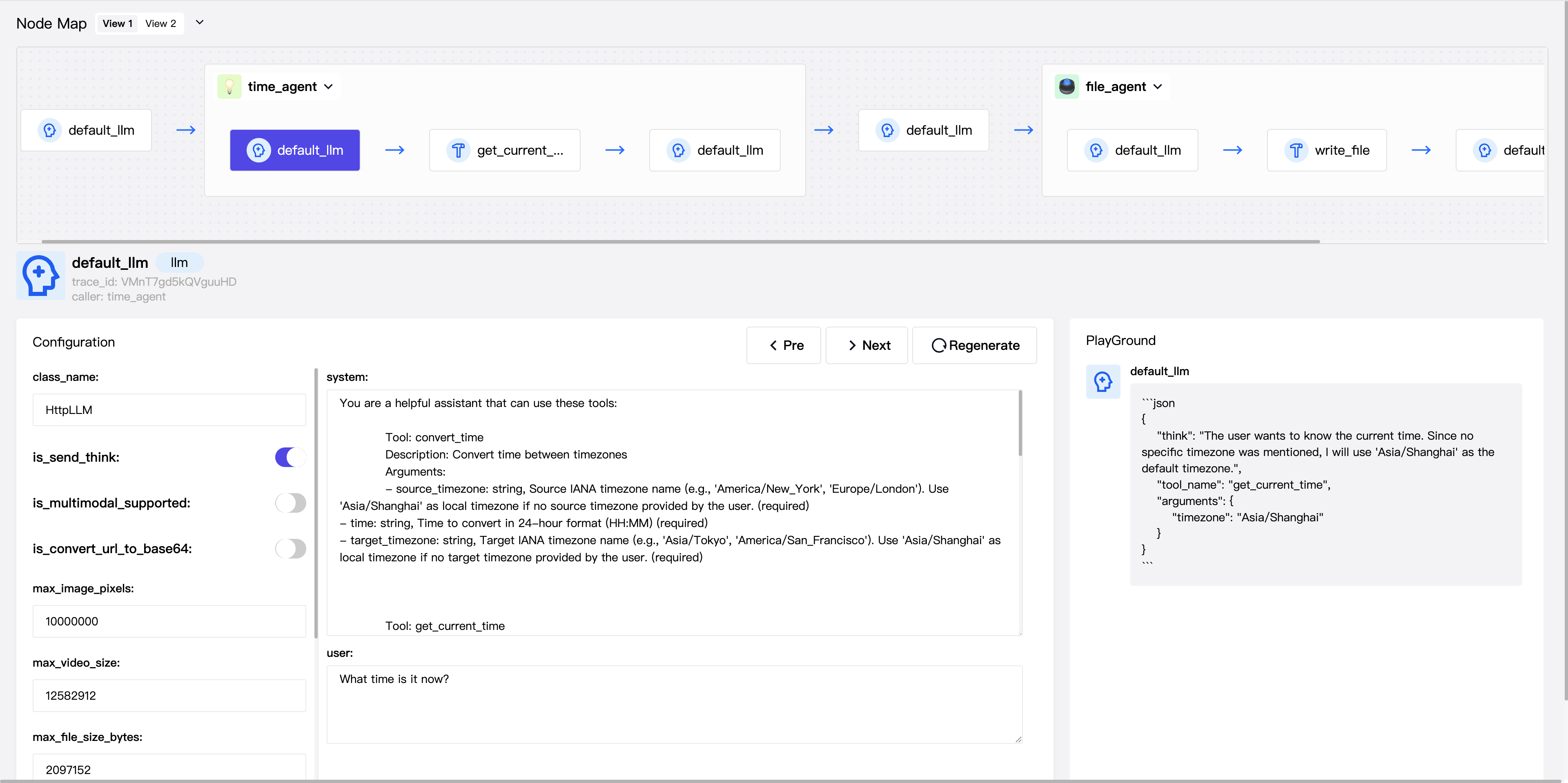}
\caption{
OxyGent can automatically generate traceable decision graphs in real time. It also supports configuration modification and regeneration of any intermediate node, making the MAS more transparent and operable.
}
\vspace{-3mm}
\label{oxy_monitor_appendix}
\end{figure*}

\textbf{Cognitive Transparency.} To address the black-box nature of multi-agent systems, OxyGent provides end-to-end visibility from high-level strategies to atomic tool operations. As shown in Figure~\ref{oxy_monitor_appendix}, the system automatically constructs traceable decision graphs that capture the rationale behind each action. This Git-like versioning of agent reasoning enables transparent collaboration and allows intermediate nodes to be regenerated under modified configurations to verify the system's adaptability.

\subsection{Details of OxyGent-GAIA}\label{appendix_gaia}

The OxyGent-GAIA is built based on DeepSeek-R1~\cite{guo2025deepseek}, GPT-4o~\cite{hurst2024gpt}, and Claude-3.5-Sonnet~\cite{anthropic2024claude}. The agentic hierarchy and tool distribution are detailed below:

\begin{itemize}
\item \textbf{Master Agent} (DeepSeek-R1): The root orchestrator that initiates and governs the entire workflow, managing the following three primary functional units:
\begin{itemize}
\item \textbf{Task Agent} (GPT-4o): Triggered first by the Master to perform high-level task decomposition and generate the overall strategic plan for the request.
\item \textbf{Coordinator Agent} (GPT-4o): Manages the specific coordination and execution phase by delegating sub-tasks to its specialized sub-agents:
\begin{itemize}
\item \textbf{Web Agent} (GPT-4o): Equipped with the \textit{SearchToolkit} (Google, Wikipedia, revisions, and archives), \textit{DocumentProcessingToolkit}, \textit{AsyncBrowserToolkit}, and \textit{VideoAnalysisToolkit} for real-time information retrieval.
\item \textbf{Document Processing Agent} (GPT-4o): Specialized in multimodal content extraction using toolkits for documents, images, audio, video, and general code execution.
\item \textbf{Reasoning Coding Agent} (GPT-4o): Focused on complex logic and data manipulation. It utilizes the \textit{CodeExecutionToolkit}, powered specifically by Claude-3.5-Sonnet, along with the \textit{ExcelToolkit} and \textit{DocumentProcessingToolkit}.
\end{itemize}
\item \textbf{Answerer Agent (GPT-4o):} Responsible for the final response synthesis. Due to GAIA's strict formatting requirements, this agent ensures precision and has the authority to reject and re-delegate the task if the reasoning evidence is deemed insufficient.
\end{itemize}
\end{itemize}

\subsection{Extended Case Studies}\label{appendix_cases}

\subsubsection{Commercial Application Patterns}

The {OxyGent} framework has established several standardized patterns for industrial MAS deployment:
\begin{itemize}
\item \textbf{Standard Operating Procedures (SOP):} Agents are assigned to specific workflow segments (\emph{e.g.}, approval, data validation). Top-level agents monitor progress while lower-level agents execute tasks, improving efficiency in standardized business processes like contract auditing.
\item \textbf{Advanced RAG Pipelines:} A multi-layered collaboration where specialized agents handle retrieval, answer generation, and quality review separately, significantly reducing hallucinations in enterprise knowledge bases.
\item \textbf{Automated Data Analytics:} {OxyGent} orchestrates agents to perform end-to-end data collection, cleaning, modeling, and visualization, streamlining the generation of automated business reports and anomaly detection.
\item \textbf{Cross-System Tool Orchestration:} In complex DevOps and office automation scenarios, agents dynamically select and invoke external APIs, maintaining system state across multiple platforms.
\end{itemize}

\subsubsection{Community and Developer Feedback}
Open-source community feedback highlights the extensibility of {OxyGent} across diverse niche domains:
\begin{itemize}
\item \textbf{Natural Language to SQL (NL2SQL):} Developers have utilized {OxyGent} to build multi-agent clusters that convert natural language queries into complex SQL statements with multi-dimensional consistency verification.
\item \textbf{Low-Code Automation:} Integration with tools like \textit{JoyCode} has enabled users to perform web searches and file operations via natural language.
\item \textbf{Database Optimization:} Autonomous agents built on {OxyGent} are currently used to diagnose ``slow SQL'' queries and generate governance suggestions to improve system performance.
\item \textbf{Dynamic Flowchart Generation:} Using local API calls, agents can automatically generate and render core implementation logic into visual flowcharts for better developer communication.
\end{itemize}

\end{document}